\documentclass[10pt,letterpaper,twocolumn]{article}

\usepackage[margin=2cm]{geometry}

\usepackage{graphicx}
\usepackage{array}
\usepackage{subfig}
\usepackage{algorithm,algorithmic}
\usepackage{color}
\usepackage{url}
\usepackage{hyperref}
\usepackage{listings}
\usepackage[none]{hyphenat}

\begin{document}

\title{Introducing ReQuEST: an Open Platform for Reproducible\\and Quality-Efficient Systems-ML Tournaments}

\author{Thierry Moreau$^{1}$, Anton Lokhmotov$^{2}$, Grigori Fursin$^{3}$ \\
\\
$^1$~University of Washington, USA ; $^2$~dividiti, UK ; $^3$~cTuning foundation, France\\
}

\date{December, 2017}





\maketitle

\begin{abstract}
 Co-designing efficient machine learning based systems across
the whole hardware/software stack to trade off speed, accuracy, energy
and costs is becoming extremely complex and time consuming.
Researchers often struggle to evaluate and compare different
published works across rapidly evolving software
frameworks, heterogeneous hardware platforms, compilers,
libraries, algorithms, data sets, models, and environments.

We~\footnote{ReQuEST organizers (A-Z): 
Luis Ceze, University of Washington (USA),
Natalie Enright Jerger (University of Toronto, Canada),
Babak Falsafi (EPFL, Switzerland),
Grigori Fursin, (cTuning foundation, France),
Anton Lokhmotov, (dividiti, UK),
Thierry Moreau, (University of Washington, USA),
Adrian Sampson, (Cornell University, USA)
Phillip Stanley Marbell, (University of Cambridge, UK)}
present our community effort to develop an open co-design tournament 
platform with an online public scoreboard.
It will gradually incorporate best research practices 
while providing a common way for multidisciplinary researchers
to optimize and compare the quality vs. efficiency Pareto
optimality of various workloads on diverse and complete
hardware/software systems.
We want to leverage the open-source Collective Knowledge framework
and the ACM artifact evaluation methodology to validate and share the complete 
machine learning system implementations
in a standardized, portable, and reproducible fashion.
We plan to hold regular multi-objective optimization and co-design tournaments 
for emerging workloads such as deep learning, starting with ASPLOS'18
(ACM conference on Architectural Support for Programming Languages and Operating Systems 
- the premier forum for multidisciplinary systems research spanning computer architecture 
and hardware, programming languages and compilers, operating systems and networking)
to build a public repository of the most efficient algorithms and systems
which can be easily reused and built upon.
We will also use the feedback from participants to continue improving
our platform and common co-design methodology.

\end{abstract}



\section{Introduction} 
\label{introduction} 

   \begin{figure*}[t]
     \centering
      \includegraphics[width=4.9in]
      {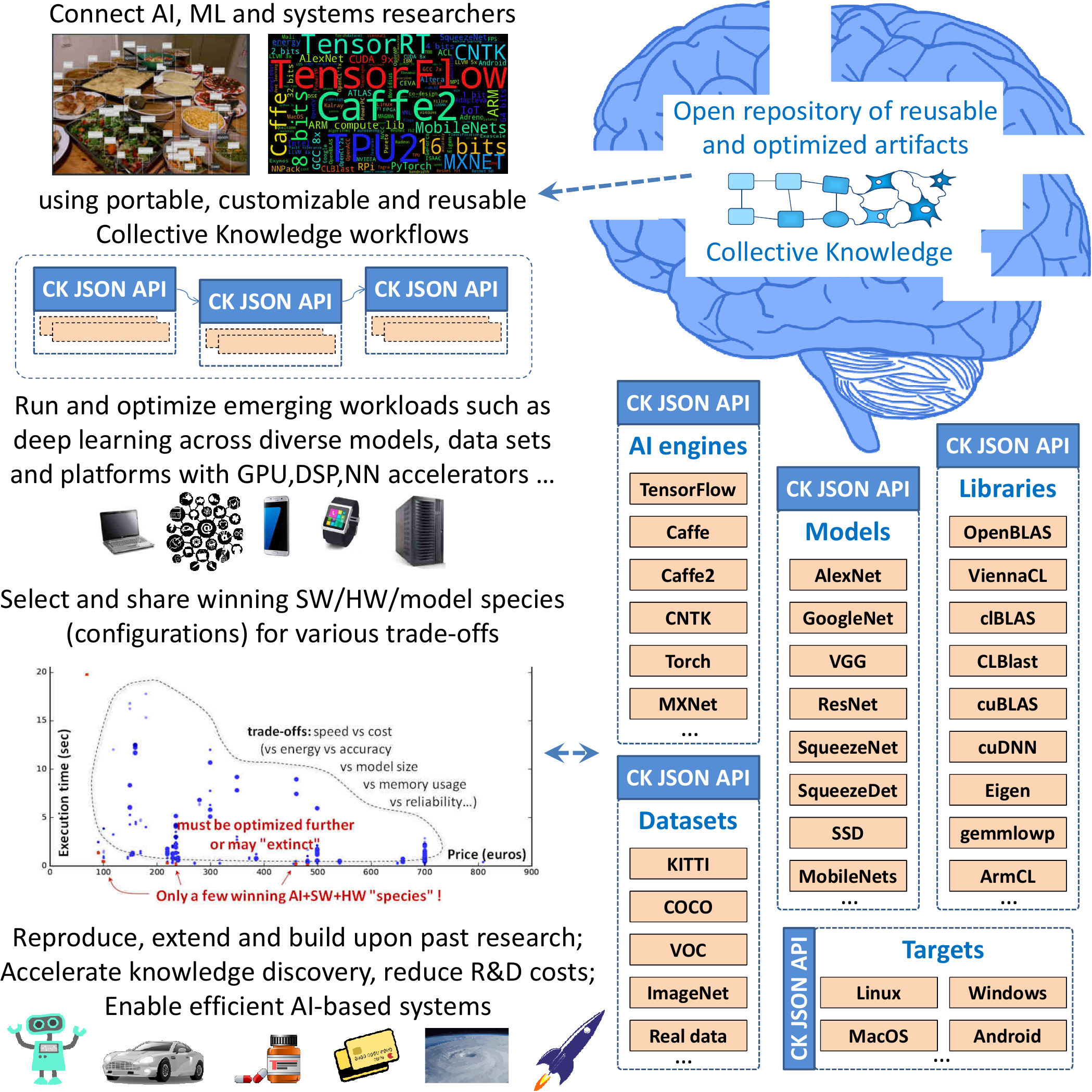} 
     \caption{
       Collective Knowledge framework as an open platform to support open tournaments 
       for software/hardware co-design of Pareto-efficient deep learning
       and other emerging workloads in terms of speed, accuracy, energy and various costs.
     }
     \vspace{-1em}
     \label{fig:reproducible-tournaments}
   \end{figure*}

Machine learning has undergone a rapid pace of progress 
over the recent years. Rarely in the scientific community have
we witnessed such a concerted effort from various communities
(machine learning, systems, hardware, security, programming
languages etc.) in improving the performance, accuracy,
robustness and cost of machine learning based systems.

However, implementing such systems for a given problem 
(for example, deep learning algorithm for ImageNet classification), 
one has to navigate a multitude of design decisions: 
what network architecture to deploy and how to customize it (ResNet vs.\ MobileNet), 
what framework to use (MXNet vs. TensorFlow), 
what libraries and which optimizations to employ (MKL vs.\ OpenBLAS),
which is generally a consequence of the target
hardware platform (Intel Xeon + NVIDIA GPU vs.\ ARM-based mobile SoC). 
On top of these implementation decisions,
platform-specific decisions may affect the performance and
overall experience in deploying the system in question:
details such as operating system, kernel version, framework
and library versions or dependencies, and custom optimizations.
As a result a given system implementation (for example, ResNet
on TensorFlow on a Intel + NVIDIA hardware system) can have
many incarnations, some of which may have drastically
different performance results. 
Furthermore, as more papers are being published, it also becomes 
challenging to reproduce, reuse, build on top of, and perform 
fair comparisons of numerous machine learning techniques 
across rapidly evolving systems.

As multiple communities tackle the same challenges 
of making machine learning systems 
faster, cheaper, smaller, more accurate, and more energy efficient 
across diverse platforms from IoT to data centers, 
we need a platform to automate and perform apples-to-apples comparisons 
of different approaches that aim to achieve the same goal. 
For example, how does an approximate analog accelerator
for deep learning compare to algorithmic simplifications
on off-the-shelf hardware in terms of accuracy vs.\ efficiency
Pareto optimality?

We therefore propose the Reproducible Quality-Efficient Systems
Tournament (ReQuEST) as a community-driven platform 
for reproducible, comparable, and multi-objective 
optimization of emerging workloads
~\cite{request}.
We plan to host ReQuEST as a bi-annual workshop alternating
between systems and machine learning communities.
Its first incarnation will take place at ASPLOS in March 2018
and will focus solely on optimizing inference on real systems 
to test our platform and use the feedback from participants and
an industrial board
to improve it.

We detail in the next sections how its objectives and
execution differentiate ReQuEST from other existing workshop-based
competitions.

\section{Main goals}

\textbf{Summary:} ReQuEST is aimed at providing a scalable tournament framework, 
a common experimental methodology and an open repository for continuous evaluation 
and optimization of the quality vs.\ efficiency Pareto optimality of a wide range 
of real-world applications, libraries, and models across the whole 
hardware/software stack on complete platforms as conceptually shown 
in Figure~\ref{fig:reproducible-tournaments}.

\textbf{Tournament framework goals:} we want to promote reproducibility 
of experimental results and reusability/customization of systems research artifacts 
by standardizing evaluation methodologies and facilitating the deployment 
of efficient solutions on heterogeneous platforms. 
For that reason, packaging artifacts and experimental results requires a bit more
involvement than sharing some CSV files or checking out a given GitHub repository. 
%
   \begin{figure*}[!htbp]
     \centering
      \includegraphics[width=5.2in]
      {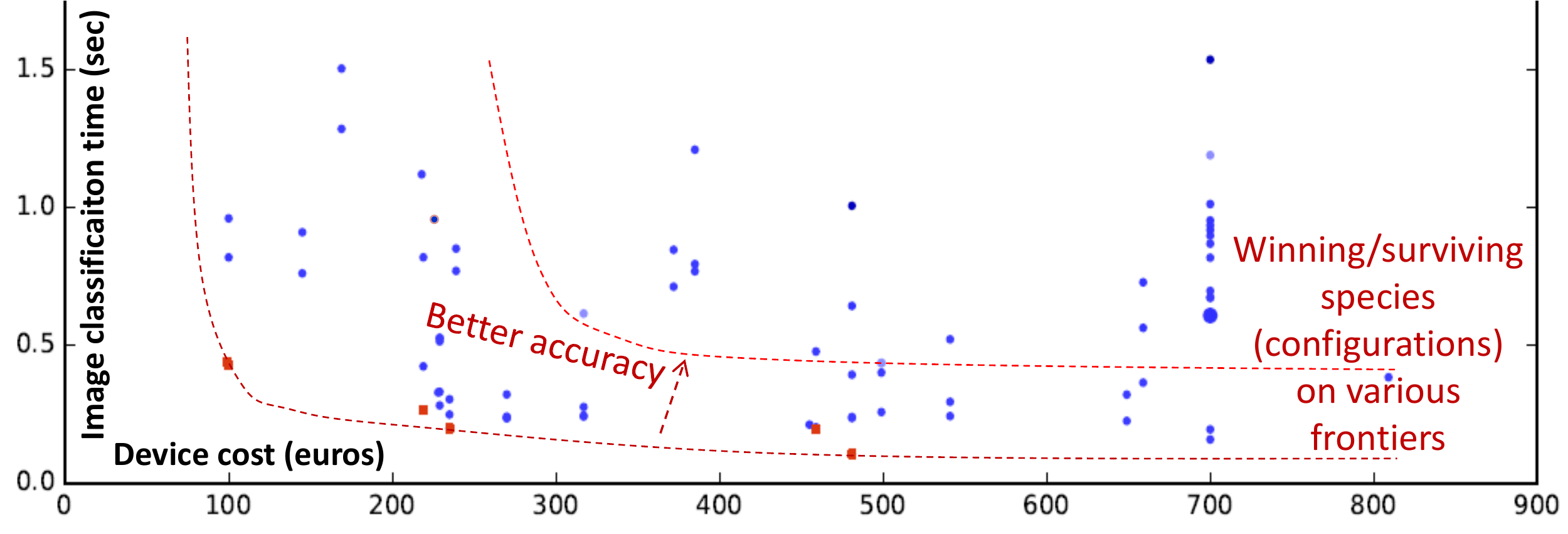} 
     \caption{
      An example of a live Collective Knowledge scoreboard to crowd-benchmark
      inference in terms of speed and platform cost 
      across diverse deep learning frameworks, models, data sets, and 
      Android devices provided by volunteers. Red dots 
      are associated with the winning workflows (model/software/hardware).
     }
     \vspace{-1em}
     \label{fig:dnn-crowdtuning-example}
   \end{figure*}

That is why we build our competition on top of an open-source 
and portable workflow framework (Collective Knowledge or CK~\cite{ck-date16})
and a standard ACM artifact evaluation methodology~\cite{ctuning-ae} 
from premier ACM systems conferences (CGO, PPoPP, PACT, SuperComputing) 
to provide unified evaluation and a live scoreboard of submissions
as demonstrated in Figure~\ref{fig:dnn-crowdtuning-example}. 

CK is a Python wrapper framework to share artifacts and workflows 
as customizable and reusable plugins with a common JSON API and meta description,
and adaptable to a user platform with Linux, Windows, MacOS and Android.
For example, it has already been used and extended in a number of academic and industrial projects
to automate and crowdsource benchmarking and multi-objective optimization of deep learning
across diverse platforms, environments, and data sets.
Figure~\ref{fig:dnn-crowdtuning-example} shows a proof-of-concept example of a live scoreboard 
powered by CK to collaboratively benchmark inference (speed vs.\ platform cost) 
across diverse deep learning frameworks (TensorFlow, Caffe, MXNet, etc.), 
models (AlexNet, GoogleNet, SqueezeNet, ResNet, etc.), real user data sets, and mobile devices 
provided by volunteers (see the latest results at \href{http://cknowledge.org/repo}{cKnowledge.org/repo}).

\textbf{Metrics and Pareto-optimality goals:} we want to stress
quality-awareness to the architecture/compilers/systems community,
and resource-awareness to the applications/algorithms community and
end-users. The submissions and their evaluation metrics will
be maintained in a public repository that includes a live
scoreboard. 

Specific attention will be brought to submissions 
close to a Pareto frontier in a multi-dimensional 
space of accuracy, execution time, power/energy consumption, 
hardware/code/model footprint, monetary costs etc.

\textbf{Application goals:} in the long term, we will
cover a comprehensive suite of workloads, datasets and models
covering applications domains that are most relevant
to machine learning and systems researchers. 
This suite will continue evolving according 
to feedback and contributions from the academia and industry.
All artifacts from this suite can be automatically plugged 
in to the ReQuEST competition workflows to simplify and automate 
experimentation.

\textbf{Complete platforms goals:} we aim to cover
a comprehensive set of hardware systems from data-centers down
to sensory nodes, incorporating various forms of processors
including GPUs, DSPs, FPGAs, neuromorphic and even analogue
accelerators in the long term.

\section{Future work}

Our goal is to bring multi-disciplinary researchers to 
\begin{enumerate}
  \item release research artifacts of their on-going or accomplished research,
standardize evaluation workflows, and facilitate deployment
and tech transfer of state-of-the-art research,
  \item foster exploration of quality-efficiency trade-offs, and 
  \item create a discussion ground to steer the community towards new
applications, frameworks, and hardware platforms.
\end{enumerate}

We want to set a coherent research roadmap for
researchers by hosting bi-annual tournaments complemented with
panel discussions from both academia and industry. We hope
that as participation increases, the coverage of problems
(vision, speech and even beyond machine learning) and
platforms (novel hardware accelerators, SoCs, and even exotic
hardware such as analog, neuromorphic, stochasticm, quantum) will increase.

ReQuEST is organized as a bi-annual workshop, alternating
between systems-oriented and machine learning-oriented
conferences. The first ReQuEST workshop will be co-located
with ASPLOS in March 2018 (ACM conference on Architectural Support for Programming Languages and Operating Systems 
- the premier forum for multidisciplinary systems research spanning computer architecture 
and hardware, programming languages and compilers, operating systems and networking).
The workshop will aim to present
artifacts submitted by participants, along with
a multi-objective scoreboard, where quality-efficient
implementations will be rewarded. 
The submissions will be validated by an artifact evaluation committee, 
and participants will have the chance to get an artifact paper
published as ACM proceedings. 

In addition we wish to nurture
a discussion ground for artifact evaluation in multidisciplinary
research, gathering perspectives from machine learning,
systems, compilers and architecture experts. 
We will use this discussion to continuously improve and extend 
functionality of our tournament platform.
For example, we plan to gradually standardize the API and meta description of all artifacts 
and machine learning workflows with the help of the community, 
provide architectural simulators and simulator-based evaluations, 
cover low-level optimizations, expose more metrics, and so on. 

Finally, an industrial panel composed of research-representatives from
prominent software and hardware companies will discuss how
tech-transfer can be facilitated between academia and
industry, and will help craft the roadmap for the ReQuEST
workshops by suggesting new datasets, workloads, metrics, and
hardware platforms.


\bibliographystyle{abbrv}
\bibliography{paper}

\newcommand{\noop}[1]{}
\begin{thebibliography}{1}

\bibitem{ctuning-ae}
{Artifact Evaluation} for computer systems conferences.
\newblock \url{http://cTuning.org/ae}, 2014--present.

\bibitem{request}
{ReQuEST}: open tournaments on collaborative, reproducible and pareto-efficient
  software/hardware co-design of emerging workloads using the collective
  knowledge technology.
\newblock \url{http://cKnowledge.org/request}, 2017--present.

\bibitem{ck-date16}
G.~Fursin, A.~Lokhmotov, and E.~Plowman.
\newblock {Collective Knowledge}: towards {R\&D} sustainability.
\newblock In {\em Proceedings of the Conference on Design, Automation and Test
  in Europe (DATE'16)}, March 2016.

\end{thebibliography}

\end{document}